\documentclass[letterpaper]{article} 
\usepackage{aaai23}  
\usepackage{times}  
\usepackage{helvet}  
\usepackage{courier}  
\usepackage[hyphens]{url}  
\usepackage{graphicx} 
\usepackage{natbib}  
\usepackage{caption} 
\usepackage{algorithm}
\usepackage{algorithmic}

\usepackage{newfloat}
\usepackage{listings}
\DeclareCaptionStyle{ruled}{labelfont=normalfont,labelsep=colon,strut=off} 
\floatstyle{ruled}
\newfloat{listing}{tb}{lst}{}
\floatname{listing}{Listing}
\usepackage[utf8]{inputenc} 
\usepackage{booktabs}       
\usepackage{amsmath}
\usepackage{amsfonts}       
\usepackage{nicefrac}       
\usepackage{subcaption}
\usepackage{xcolor}         

\usepackage{pdfpages}

\definecolor{cerulean}{rgb}{0.0, 0.48, 0.65}
\definecolor{gray}{cmyk}{0.0, 0.0, 0.0, 0.65}

\newcommand\smalleq{\mkern1.5mu{=}\mkern1.5mu}

\title{HyperJump: Accelerating HyperBand via Risk Modelling}

\author {
    Pedro Mendes\textsuperscript{\rm 1,2}, 
    Maria Casimiro\textsuperscript{\rm 1,2},
    Paolo Romano\textsuperscript{\rm 2}, 
    David Garlan\textsuperscript{\rm 1} 
}
\affiliations {
    \textsuperscript{\rm 1} Software and Societal Systems Department, Carnegie Mellon University\\
    \textsuperscript{\rm 2} INESC-ID and Instituto Superior Técnico, Universidade de Lisboa\\
    \{pgmendes, mdaloura, dg4d\}@andrew.cmu.edu,
    romano@inesc-id.pt
}

\begin{document}

\maketitle

\begin{abstract}
In the literature on hyper-parameter tuning, a number of recent solutions rely on low-fidelity observations (e.g., training  with sub-sampled datasets) in order to efficiently identify  promising configurations to be then tested via high-fidelity observations (e.g., using the full dataset).  Among these, HyperBand is arguably one of the most popular solutions, due to its efficiency and theoretically provable robustness.

In this work, we introduce HyperJump, a new approach that builds on HyperBand's robust search strategy and complements it with novel model-based risk analysis techniques that accelerate the search by skipping the evaluation of low risk configurations, i.e., configurations that are likely to be eventually discarded by HyperBand.
We evaluate HyperJump on a suite of hyper-parameter optimization problems and show that it provides over one-order of magnitude speed-ups, both in sequential and parallel deployments, on a variety of deep-learning, kernel-based  learning and neural architectural search problems when compared to HyperBand and to several  state-of-the-art optimizers.

\end{abstract}

\section{Introduction}

\label{sec:intro}

Hyper-parameter tuning is a crucial phase to optimize the performance of machine learning (ML) models, which is notoriously expensive given that it typically implies repeatedly training models over large data sets. State-of-the-art solutions address this issue by exploiting cheap, low-fidelity models (e.g., trained with a fraction of the available data) to extrapolate the quality of fully trained models.



HyperBand~\cite{hyperband}, henceforth referred to as HB,  is one of the most popular solutions in this area. HB is based upon a randomized search procedure, called Successive Halving (SH)~\cite{successiveHalving}, which operates in stages of fixed ``budget'' $R$ (e.g.,  training time or training set size): at the end of  stage $i$, the best performing 1/$\eta$ configurations are selected to be evaluated in stage $i+1$, where they will be allocated $\eta\times$ larger budget (see the bottom diagram of Fig.~\ref{fig:HJ_vs_HB}). By restarting the SH procedure over multiple, so called, brackets using  different initial training budgets, HB provides theoretical guarantees of convergence to the optimum, incurring negligible computational overheads and outperforming state-of-the-art optimizers (e.g., based on Bayesian Optimization (BO) \cite{boTutorial}) that do not exploit low-fidelity observations. However, the random nature of HB also inherently limits its efficiency, as shown by recent model-based multi-fidelity approaches~\cite{bohb,fabolas}. 

We introduce HyperJump (HJ), a novel hyper-parameter optimization method that builds upon HB's robust search strategy and accelerates it via an innovative, model-based technique. The idea at the basis of HJ is to ``jump'' (i.e., skip either partially or entirely) some of HB's stages (see the top diagram of Fig.~\ref{fig:HJ_vs_HB}). 
To minimize the risks associated with jumps, while  maximizing the attainable gains  by favoring earlier jumps, HJ exploits, in a synergistic way, three new mechanisms:

\begin{figure*}
    \centering
    \includegraphics[scale=0.43]{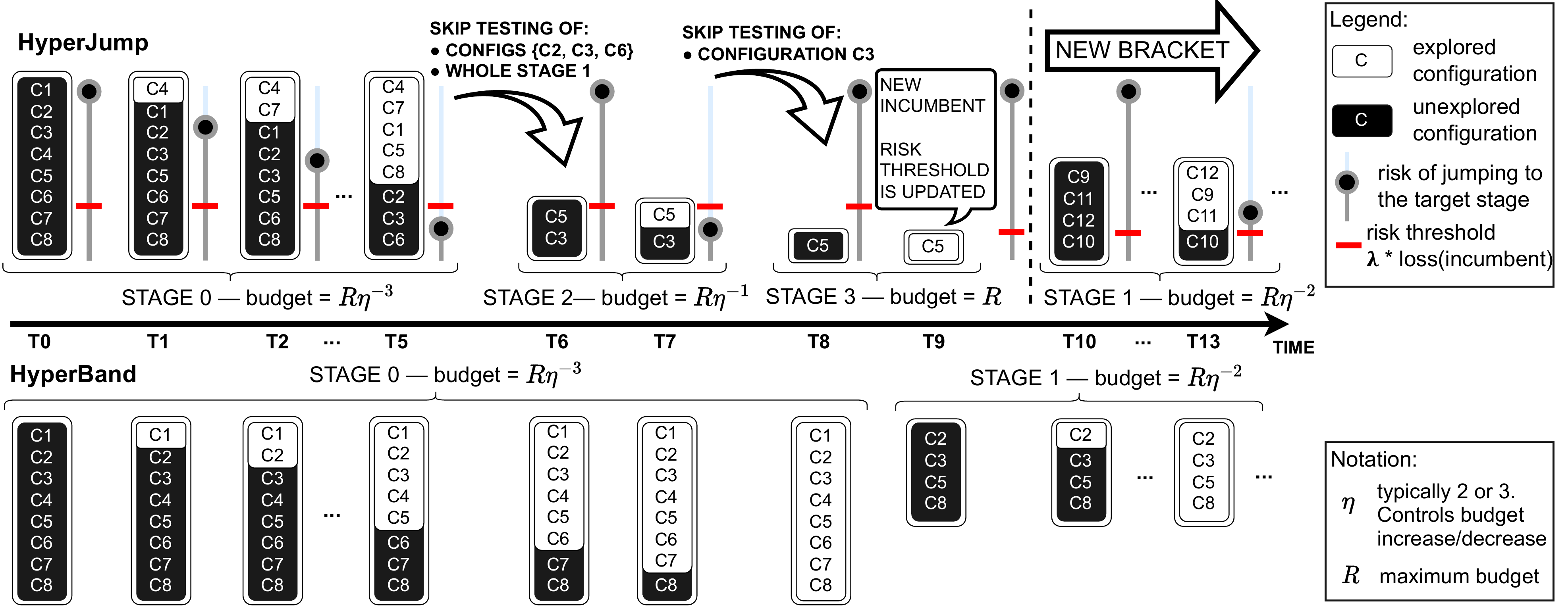}
    \caption{Search methodologies of HJ (top) and HB (bottom). The figure illustrates the 3 key mechanisms of HJ: i) skipping HB stages based on a risk model; ii) determining the order in which configurations are test so as to minimize the risk of future jumps (this is depicted in the figure through the different set of configurations explored by each approach); iii) dynamically updating the risk threshold based on the quality of the current incumbent.
    }
    \label{fig:HJ_vs_HB}
\end{figure*}




\begin{itemize}
    \vspace{-1mm}
    \item Expected Accuracy Reduction (EAR) --- a novel modelling technique to predict the risk of jumping. The EAR exploits the model's uncertainty in predicting the quality of untested configurations as a basis to estimate the expected reduction in the accuracy between   \textbf{(i)} the best configuration included in the stage reached after a jump and  \textbf{(ii)}  the best configuration  discarded due to the jump. 
    
    \vspace{-1mm}
    \item A criterion for selecting the configurations to include in the HB stage targeted by a jump, which aims to minimize the jump's risk. This is  a combinatorial problem\footnote{The number of candidate configuration sets for a jump to a stage with $k$ configurations from one with $n$ configurations is $\binom{n}{k}$.}, which we tackle via a lightweight  heuristic that has  logarithmic complexity with respect to the number of  configurations in the target stage of the jump.

    \item A method for prioritizing the testing of configurations that aims to promote future jumps, by favouring the sampling of configurations that are expected to yield the highest risk reduction for future jumps. 
\end{itemize}



We conduct an ablation study that sheds light on the contributions of the various mechanisms of HJ on its performance, and compare HJ with a number of state-of-the-art optimizers~\cite{hyperband,fabolas,Practical_BO,bohb,asha} on both hyper-parameter optimization and neural architecture search~\cite{nats} problems, for sequential and parallel deployments. We show that HJ provides up to over one-order of magnitude speed-ups on deep-learning and kernel-based learning problems. 
 


\section{Related Work}

\label{sec:rw}

Existing hyper-parameter techniques can be coarsely classified along two dimensions: \textbf{i)} whether they use model-free or model-based approaches; \textbf{ii)} whether they exploit solely high-fidelity evaluations or also multi-fidelity ones.

As already mentioned, 
HB  is arguably one of the most prominent model-free approaches. Its random nature, combined with its SH-based search algorithm, makes it not only provably robust, but also efficiently parallelizable~\cite{hyperband,bohb,asha} and, overall, very competitive  and  lightweight. 

As for the model-based approaches, recent literature on hyper-parameter optimization has been dominated by  Bayesian Optimization (BO)~\cite{boTutorial} methods. BO relies on modelling techniques (e.g., Gaussian Processes~\cite{GP_book}, Random Forests~\cite{Random_Forest} or Tree Parzen Estimator (TPE)~\cite{tpe}) to build a surrogate model of the function $f:\mathcal{X}\rightarrow \mathbb{R}$ to be optimized. The surrogate model is then used to guide the selection of the configurations to test via an \textit{acquisition function} that tackles the exploration-exploitation dilemma. A common acquisition function is the Expected Improvement (EI) \cite{ei}, which exploits information of the model's uncertainty on an untested configuration $c$ to estimate by how much $c$ is expected to improve over the current incumbent.

SMAC~\cite{smac} and MTBO~\cite{mtbo} were probably the first to  adapt the BO framework to take advantage of low-fidelity evaluations, obtained using  training sets of smaller dimensions. This idea was extended in Fabolas~\cite{fabolas}, whose 
acquisition function factors in the trade-off between the cost of testing a configuration and the knowledge it may reveal about the optimum. 
%
A related body of work~\cite{freeze-thawBO,loss_NN_gps,bo_stop,Vizier,mes-par,ucb-mf-gp,mf-hp,miso,general-mfbo} 
uses models (typically GPs) to predict the loss of neural networks as a function of both the hyper-parameters and the training iterations. Models are then used to extrapolate the full-training loss and cancel under-performing training runs. 

From a methodological perspective, the fundamental difference between HJ and this body of model-based works lies in the type of problems that are addressed by their modelling techniques. Since HJ exploits HB's search strategy, the models employed in HJ are meant to address a different problem than the above mentioned model-based approaches, namely quantifying the risk of discarding high quality configurations among the ones currently being tested in the current HB's stage. In other words, HJ's models are used to reason on a relatively small search space, i.e.,~the configurations in the current stage. Conversely, the previously mentioned approaches, which do not take advantage of HB's search algorithm, employ models to estimate configuration's quality across the whole search space.
Further, the modelling techniques used in some of these works~\cite{fabolas} require computationally expensive implementations, which can impose significant overhead especially when the cost of evaluating configurations is relatively cheap, e.g., when using low-fidelity observations (see Sec.~\ref{sec:eval}). 
%


By taking advantage of the (theoretically provable) robustness of HB, which HJ accelerates in a risk-aware fashion via model-driven techniques, we argue that HJ fuses the best of both worlds: it preserves HB's  robustness (see Sec.~\ref{sec:theo}), 
 while accelerating it by more than one order of magnitude  via the use of lightweight, yet effective, modelling techniques (as we will show experimentally in Sec.~\ref{sec:eval}).

The works more closely related to HJ are recent approaches, e.g.,~\cite{bohb,bohb_similar1,bohb_similar2} 
that extend HB with BO, or evolutionary search~\cite{dehb}, to \textit{warm start} it, i.e., to select (a fraction of) the configurations to  include in a new HB bracket. 
This mechanism is complementary to the key novel idea exploited by HJ to accelerate HB, i.e., short-cutting the SH process by skipping to later SH stages in low risk scenarios. In fact, HJ incorporates a BO-based warm starting mechanism and it would be straightforward to incorporate alternative approaches (e.g., based  on evolutionary search~\cite{dehb}). 
Also, when compared with BOHB~\cite{bohb},  which warm starts HB via BO, HJ provides more than one order of magnitude speed-ups. 

Recently, ASHA~\cite{asha} proposed an optimized parallelization strategy for HB that aims at avoiding struggling effects. In  Section~\ref{sec:parallel} we discuss how HJ's approach could be applied to ASHA. Further, our experimental study shows that HJ (despite using HB's original parallelization scheme) still signficantly outperforms ASHA.
\begin{algorithm}[t]
    \caption{Pseudo-code for a HJ bracket consisting of $S$ stages, with budget $b$ for the  initial stage.}
    \scriptsize
    \label{alg:overall}

    \begin{algorithmic}[1]
    
    
    \STATE {Set}$\langle$Config$\rangle$ $C$ = \textsc{get\_configs\_for\_bracket}() \COMMENT{Model-driven bracket warmstart (Suppl. Mat Sec. 6)} \label{alg:get_conf} 
    
    \STATE {Set}$\langle$Config$\rangle$ T = $\emptyset$; {Set}$\langle$Config$\rangle$U = $C$; \COMMENT{T and U contain the tested/untested configs, resp.} 
    \FOR [s denotes the current stage] {$s \in \{0,\ldots, S-1\}$}

    \STATE \textbf{bool} jump = \textit{false};
    
    \WHILE [Test configs. in curr. stage, or jump to a future stage] {$U \neq \emptyset$}
    
    \STATE $\langle$target, $\mathcal{S}\rangle$ = \textsc{evaluate\_jump\_risk}($s$, T, U) \COMMENT{HJ risk-analysis (Sec.~\ref{sec:risk})} \label{alg:risk}  
    
    \IF [Jump to target stage]{target $\neq s$} 
    \STATE $s$ = target; T = U = $\emptyset$; $C$ = $\mathcal{S}$ ; jump = \textit{true}; \textbf{break};
    \ELSE
    \STATE c = \textsc{next\_conf\_to\_test}(U, T, $s$); \COMMENT{Next selected config.~minimizes future risk (Sec.~\ref{sec:order})}  \label{alg:test} 
    \STATE acc = \textit{evaluate\_config}(c,$b\eta^s$) \COMMENT{Measure  config. c with budget $b\eta^s$ and return its accuracy}
    \STATE T = T $\cup~\{c\}$;   U = U $\setminus$ \{c\}
    \STATE \textit{update\_model}($\langle c,b\eta^s,acc\rangle$)
    
    \ENDIF
    \ENDWHILE
    \IF [Use HB's policy if HJ did not trigger a jump by the end of the stage]{$\neg$jump} 
    \STATE U = $C$ = topK($C$, $|C|\eta^{-1}$) \COMMENT{test top $1/|\eta|$\% configs. in next stage}
    \ENDIF
    \ENDFOR
    \end{algorithmic}
\end{algorithm}
\setlength{\textfloatsep}{2mm}

\section{HyperJump}

\label{sec:hyperjump}

As already mentioned, an HB bracket is composed of up to $S_{max}$ stages, where $S_{max}=\lfloor log_{\eta}(R)\rfloor$. $R$ is the maximum ``budget''  allocated to the evaluation of a hyper-parameter configuration and $\eta$ is an exponential factor (typically 2 or 3) that controls the increase/decrease of the allocated budget/number of tested configurations in two consecutive stages of the same bracket. 

The pseudo-code in Alg.~\ref{alg:overall} overviews  the various mechanisms employed by HJ to accelerate a single HB bracket composed of $S<S_{max}$ stages. The bracket's initial stage allocates a budget $b$, where $b=R\eta^{-(S-1)}$, as prescribed by HB. 
HJ leverages two main mechanisms to accelerate the execution of a HB bracket, which are encapsulated in the functions \textsc{evaluate\_jump\_risk} and \textsc{next\_config\_to\_test}. 

\textsc{evaluate\_jump\_risk} is executed within HJ's inner loop (Alg.~\ref{alg:overall}, line~\ref{alg:risk}), to decide whether to stop testing configurations in the current stage and jump to a later stage. This function takes as input the current stage, $s$, and the configurations already and still to be tested, $T$ and $U$. It returns the pair $\langle target, \mathcal{S}\rangle$ where $target$ denotes the stage to jump to (in which case $target \neq s$) and $\mathcal{S}$ the selected configurations for the target stage. We detail this function  in Sec.~\ref{sec:risk}.

If the risk of jumping is deemed too high, HJ continues testing configurations in the current stage. Unlike HB, which uses a random order of exploration, HJ prioritizes the order of exploration via the \textsc{next\_conf\_to\_test} function (Alg.~\ref{alg:overall}, line~\ref{alg:test}). This function  seeks to identify a configuration whose evaluation will lead to a large reduction of the risk of jumping, so as to favour earlier jumps and enhance the  efficiency of HJ. We discuss how we implement this function in Sec.~\ref{sec:order}. 
After testing configuration $c$ in budget $b\eta^s$ and measuring its accuracy $acc$, we update the models used to predict the accuracy of untested configurations (with different budgets). 

Finally,  \textsc{get\_configs\_for\_bracket} (Alg.~\ref{alg:overall}, line~\ref{alg:get_conf}) encapsulates the model-driven warm starting procedure, which selects the configurations to include in a new bracket. As mentioned, this idea was already explored in prior works, e.g., BOHB~\cite{bohb}, so we detail its implementation in the supplemental material~\cite{HJ-arxiv}, along with the description of other previously published optimizations that we incorporated in HJ (e.g., resuming training of configurations previously on smaller budgets~\cite{freeze-thawBO}). The supplemental material includes also an analysis of HJ's algorithmic complexity  (Sec.~7). 

\subsection{Deciding whether to jump}

\label{sec:risk}

We address the problem of deciding whether to adopt HB's default policy or skip some, or even all, of the future stages of the current bracket by decomposing it into three simpler sub-problems:
\textbf{1)} modelling the risk of  jumping from the current stage to the next stage while retaining an arbitrary subset $\mathcal{S}$ of the configurations $C$ in the current stage;
\textbf{2)} identifying ``good'' candidates for the subset of  configurations to  retain after a jump from stage $s$ to stage $s+1$, i.e.,  configurations that, if included in the target stage of the jump, reduce the risk of jumping;
\textbf{3)} generalizing the risk modelling to jumps that skip  an arbitrary number of stages.
Next, we discuss how we address each sub-problem.

Firstly, HJ's risk analysis methodology  relies on GP-based models to estimate the accuracy of a configuration $c$ in a given budget.
As in recent works, e.g.,~\cite{fabolas,klein2020model,trimtuner}, we include in the feature space of the GP models not only the hyper-parameters' space, but also the budget (so as to enable inter-budget extrapolation). Further, analogously to, e.g.,~\cite{fabolas,trimtuner}, we employ a custom kernel that encodes the expectation that the loss function has an exponential decay with larger budgets along with a generic Matérn 5/2 kernel that is used to capture relations among the hyper-parameters.






\paragraph{1. Modelling the risk of 1-hop jumps.}
Let us start by discussing how we model the risk of jumping from a source (current) stage $s$ to a target stage $t$ that is the immediate successor of $s$ ($t\!\!=\!\!s\!+\!1$). 
Consider $C$ the configurations in $s$; $T$ and $U$ the tested/untested configurations in $s$, resp., and $\mathcal{S}$ and $\mathcal{D}$ the subset of configurations in $C$ that are selected/discarded, resp., when jumping to stage $t$. 
Our modelling approach is based on the observation that  short-cutting  HB's search process and jumping to the next stage exposes the risk of discarding the configuration that achieves maximum accuracy in the current stage (and that may turn out to improve the current  incumbent when tested in full-budget). This risk can be modelled as the difference between two random variables defined as the maximum accuracy of the configurations in the set of discarded $\mathcal{D}$ and selected $\mathcal{S}$ configurations in stage $s$ and budget $b\eta^s$, resp.:

$$\mathcal{A}^\mathcal{D}_{s} =\underset{c \in \mathcal{D}}{\text{max}} ~A(c, b\eta^s), ~~~\mathcal{A}^\mathcal{S}_{s} = \underset{c \in \mathcal{S}}{\text{max}} ~A(c, b\eta^s)$$
having denoted with $A(c_i, b\eta^s)$ the accuracy of a configuration $c_i$ using budget $b\eta^s$.
From a mathematical standpoint, we  only require $\mathcal{A}$ to be finite, so that the maximum and difference operators are defined. 
So, one may use arbitrary accuracy metrics as long as they  match this assumption, e.g., unbounded but finite metrics like negative log likelihood.

One can then quantify the ``absolute'' risk of a jump from stage $s$ to stage $s+1$, which we call Expected Accuracy Reduction (EAR) (Eq.~\eqref{eq:eal}), as the expected value of the difference between these two variables, restricted to the scenarios in which configurations with higher accuracy are discarded due to jumping (i.e., $\mathcal{A}^\mathcal{D}-\mathcal{A}^\mathcal{S}>0$):
%
\begin{footnotesize}
\begin{equation}
\begin{aligned}
 EAR_{s}^{s+1}(\mathcal{D},\mathcal{S}) &\smalleq 
\int_{-\infty}^{+\infty}  \!\! P \!\left( 
\mathcal{A}^\mathcal{D}_{s} \!- \!\mathcal{A}^\mathcal{S}_{s}  
\smalleq x \right)  \text{max}\{\mathcal{A}^\mathcal{D}_{s} \!- \! \mathcal{A}^\mathcal{S}_{s},0\} dx  \\& \smalleq
\int_{0}^{+\infty} \!\!  x P\left( 
\mathcal{A}^\mathcal{D}_{s}  \!- \! \mathcal{A}^\mathcal{S}_{s}  
 \smalleq x \right) dx
\end{aligned}
\label{eq:eal}
\end{equation}
\end{footnotesize}
The EAR is computed as follows. The configurations in $\mathcal{D}$ and $\mathcal{S}$ are either untested or already tested. In the former case, we model their accuracy via a Gaussian distribution (given by the GP predictors); in the latter case, we model their  accuracy either as a Dirac 
function (assuming noise-free measurements, which is HJ's default policy) or via a Gaussian distribution (whose variance can be used to model  noisy measurements). In any case, the PDF and CDF of $\mathcal{A}^\mathcal{D}$ and $\mathcal{A}^\mathcal{S}$ can be computed in closed form. Yet, computing the difference between these two random variables requires solving a convolution that cannot be determined analytically.  Fortunately, both this convolution and the outer integral in Eq.~\eqref{eq:eal}  can be  computed in a few msecs using open source numerical libraries (Sec.~\ref{exp_res}). Additional details on the EAR's computation  are provided in the supplemental material.

Next, we introduce the rEAR (relative EAR), which is obtained by normalizing the EAR by the loss of the current incumbent, noted $l^{*}$: 
$rEAR_{s}^{s+1}(\mathcal{D},\mathcal{S})=
{EAR_{s}^{s+1}(\mathcal{D},\mathcal{S})}/{l^*}$.
The rEAR  estimates the ``relative'' risk of a jump 
and can be interpreted as the percentage of the maximum potential for improvement that is expected to be sacrificed by a jump.  
%
%
%
In HJ, we consider a jump ``safe" if its corresponding rEAR is below a threshold $\lambda$, whose default value we set to $10\%$. As we show in the supplemental material,
in practical settings,  HJ has robust performances for a large range of (reasonable) values of $\lambda$. 
One advantage of using the rEAR, instead of the EAR as risk metric is that the rEAR allows for 
adapting the risk propensity of HJ's logic, making  HJ progressively less risk prone as the optimization process evolves and new incumbents are found. 


\begin{algorithm}[t]
    \caption{Pseudo-code of the logic used to determine the sets of configurations to consider when jumping from stage $s$ to stage $s+1$ (function \textsc{get\_candidates\_for\_$\mathcal{S}$}())
    }
    \label{alg:candidateSets}
    \scriptsize

    \begin{algorithmic}[1]
         \STATE Set$\langle$Set$\langle$Config$\rangle\rangle$  \textsc{\textsc{get\_candidates\_for\_$\mathcal{S}$}}(Set$\langle$Config$\rangle$ Tested,     Set$\langle$Config$\rangle$ Untested)

        \STATE Set$\langle$Set$\langle$Config$\rangle\rangle O \gets \emptyset$  \COMMENT{Set of selected safest sets to be returned}
        \STATE Set$\langle$Config$\rangle$ $C \gets$ Tested $\cup$ Untested \COMMENT{$C$ stores all the  configs in current stage}
        \STATE Set$\langle$Config$\rangle$ $\mathcal{K} \gets C.sortByAccuracy().getTop(|C|/\eta)$  \label{alg:sortAcc}
        \STATE \COMMENT{$\mathcal{K}$ stores the configs.~with best (pred. or measured) accuracy }
        \STATE Set$\langle$Config$\rangle$ $E \gets C \setminus \mathcal{K}$
        \STATE \COMMENT{$E$ stores the configs.~with worse (pred. or measured) accuracy}

        \STATE $O.add(\mathcal{K})$ \COMMENT{Set $\mathcal{K}$ is one of the $1 + 2 \lfloor log_{\eta}|\mathcal{K}|\rfloor$ recommended sets}
        
        \STATE \COMMENT{Generate sets based on accuracy}
        \FOR {$i$ in $[ 1\leq i\leq\lfloor log_{\eta}|\mathcal{K}| \rfloor$} \label{alg:prem_Acc_init}
            \STATE Set $X \gets clone(\mathcal{K})$
            \STATE  \COMMENT{Remove from $K$ the $\mathcal{|K|}/\eta^i$ configurations with worse accuracy...}
            \STATE $X.removeBottom(\mathcal{|K|}/\eta^{i})$ 
            \STATE \COMMENT{...and add  the $\mathcal{|K|}/\eta^i$ configurations with best accuracy from $C\setminus \mathcal{K}$}
            \STATE $X.add(E.getTop(\mathcal{|K|}/\eta^{i}))$ 
            \STATE $O.add(X)$
        \ENDFOR \label{alg:prem_Acc_end}
        
        \STATE $\mathcal{K} \gets \mathcal{K}.sortByLCB()$  \COMMENT{sort configs in $\mathcal{K}$ by LCB }
        \STATE $E \gets E.sortByUCB()$  \COMMENT{sort configs in $E$ by UCB}
        
        \STATE \COMMENT{Generate sets based on lower and upper confidence bounds}
        \FOR{$i$ in $[ 1\leq i\leq\lfloor log_{\eta}|\mathcal{K}| \rfloor$} \label{alg:prem_UCB_init}
            \STATE Set $X \gets clone(\mathcal{K})$
            \STATE  \COMMENT{Remove from $K$ the $|K|/\eta^i$ configurations with worse LCB}
            \STATE $X.removeBottom(\mathcal{|K|}/\eta^i)$
            \STATE \COMMENT{...and add  the $|\mathcal{K}|/\eta^i$ configurations with best UCB from $C\setminus \mathcal{K}$}
            \STATE $X.add(E.getTop(| \mathcal{K}|/\eta^i))$ 
            \STATE $O.add(X)$
        \ENDFOR \label{alg:prem_UCB_end}
        
        \STATE  \textbf{return } $O$
    \end{algorithmic}
\end{algorithm}

\paragraph{2. Identifying the safest set of configurations for a jump.}
%
Determining the safest subset of configurations to include when jumping to the next stage via naive, enumerative methods would have prohibitive costs, as it would require evaluating the rEAR for all possible subsets $\mathcal{S}$ of size $|C|/\eta$ of the configurations in the current stage $s$. For instance, assuming $\eta \!=\! 3$, $|C| \!=\! 81$ and that less than half of the configurations in $C$ were tested, the  number of distinct target sets for a jump of a single stage is 
 $\binom{81}{27}\approx$~2E21.

We tackle this problem by introducing an efficient, model-driven heuristic that recommends a total of 1+2$\lfloor$log$_{\eta}|\mathcal{S}|\rfloor$ candidates for $\mathcal{S}$ (see Alg.~\ref{alg:candidateSets}). The first candidate set evaluated by HJ, denoted $\mathcal{K}$, is obtained by considering the top $|\mathcal{S}|$ configurations, ranked based on their actual or predicted accuracy depending on whether they have already been tested or not (line~\ref{alg:sortAcc}). 

Next, using $\mathcal{K}$  as a template, we generate $\lfloor$log$_{\eta}|\mathcal{S}|\rfloor$ alternative  sets by replacing the worst $|\mathcal{S}|/\eta^i$ (1$\leq i\leq\lfloor $log$_{\eta}|\mathcal{S}| \rfloor$) configurations in $\mathcal{K}$  with the next best configurations in $C\setminus \mathcal{K}$, ranked based on their (predicted/measured) accuracy (lines~\ref{alg:prem_Acc_init}-\ref{alg:prem_Acc_end}). 
Next (lines~\ref{alg:prem_UCB_init}-\ref{alg:prem_UCB_end}), we generate $\lfloor$log$_{\eta}|\mathcal{S}|\rfloor$ alternative candidate sets by  exploiting  model uncertainty as follows: we identify the worst $|\mathcal{S}|/\eta^i$ configurations in $\mathcal{K}$,  ranked according to their lower confidence bound, and replace them with the configurations in  $C\setminus \mathcal{K}$ that have the highest confidence (we use a confidence bound of 90\%).  
Intuitively, this way we remove from the reference set $\mathcal{K}$ the configurations that are likely to have lower accuracy if the model overestimated their mean. These are replaced by the configurations that, although having lower (average) predicted accuracy, 
are prone to achieve high accuracy, given the model's uncertainty. Refer to the supplemental material for diagrams illustrating the behavior of this algorithm.

\paragraph{3. Generalizing to multi-hop jumps.}
Alg.~\ref{alg:evalRisk} shows how to compute the rEAR of a jump that skips $j\!>\!1$ stages from the current stage $s$. This is done in an iterative fashion by computing the rEAR for jumps from stage $s+i$ to stage $s+i+1$ ($i\in[1,j-1]$) and accumulating the corresponding rEARs to yield the rEAR of the jump. 

At each iteration, the candidate sets for the set $\mathcal{S}$ of configurations to be retained after the jump are obtained via the \textsc{get\_candidates\_for\_$\mathcal{S}$} function. 
Among these 1+2$\lfloor$log$_{\eta}|\mathcal{S}|\rfloor$  candidate sets, the one with minimum risk is identified. The process is repeated replacing $C$ with the candidate set that minimizes the risk of the current jump (line~\ref{alg:repl}), until $\lambda$ is exceeded, thus seeking to maximize the ``jump length'', i.e., the number of stages that can be safely skipped.
As such, the computation of the risk of a jump from a stage with $|C|$ configurations and that skips $j\!>\!1$ stages requires $\mathcal{O}(j(1+log_{\eta}|C|/~\eta))$ rEAR evaluations. This ensures the scalability of HJ's risk analysis methodology even when considering jumps that can skip a large number of stages.

\begin{algorithm}[t]
    \caption{Pseudo-code for the \textsc{Evaluate\_Jump\_Risk} function.}
    \label{alg:evalRisk}
    \scriptsize

    \begin{algorithmic}[1]
    
    
    \STATE $\langle$int s, Set$\langle$Config$\rangle\mathcal{S}\rangle$ \textsc{\textsc{Evaluate\_Jump\_Risk}}(int s, Set$\langle$Config$\rangle$ Tested, Set$\langle$Config$\rangle$ Untested)

    \STATE rEAR = 0; $\mathcal{S}$ = $\emptyset$; $C$ = Tested $\cup$ Untested.
    
    \WHILE [$S$: maximum number of stages] {s $<$ S}
    \STATE Set$\langle$Set$\langle$Config$\rangle\rangle$ candidates = \textsc{\textsc{get\_candidates\_for\_$\mathcal{S}$}}(Tested, Untested)
    \STATE minRisk = min$_{\mathcal{X}\in \text{candidates}}$ rEAR$_{s}^{s+1}(\mathcal{X}, C\setminus \mathcal{X})$
    \STATE $\mathcal{S}=\text{argmin}_{\mathcal{X}\in \text{candidates}}$ rEAR$_{s}^{s+1}(\mathcal{X}, C\setminus \mathcal{X})$ \COMMENT{$\mathcal{S}$: Set that minimizes the risk}
    
    \IF {rEAR + minRisk $ > \lambda$} 
        \STATE \textbf{return} $\langle$ s , $\mathcal{S}\rangle$ \COMMENT{Return the target stage and the set with selected configs. } 
    \ELSE [Try to extend the jump by one hop]
        \STATE rEAR += minRisk;  s++
        \STATE $C$ = Untested = $\mathcal{S}$; Tested = $\emptyset$ \label{alg:repl} 
      \ENDIF  
     \ENDWHILE
     
      \STATE  \textbf{return} $\langle$ s, $\emptyset \rangle$ \COMMENT{Jump all stages in the current bracket and start a new bracket}
    \end{algorithmic}
\end{algorithm}

\subsection{Reducing the risk of jumping by prioritizing the evaluation order of configurations}

\label{sec:order}

HJ also uses a model-based approach to determine in which order to evaluate the configurations of a stage. This mechanism aims to enhance HJ's efficiency by prioritizing the evaluation of  configurations that are expected to yield the largest reduction of risk for future jumps. 
The literature on look-ahead non-myopic BO~\cite{yue2020nonmyopic,lynceus,lookahed_bo_wilcox,Lam:2017} has already investigated several techniques aimed at predicting the impact of future exploration steps on model-driven optimizers. These approaches often impose  large computational overheads, due to the need to perform expensive ``simulations'', e.g., retraining the models to simulate alternative evaluation outcomes, and to their ambition to maximize long-term rewards (in contrast to the greedy nature of typical BO approaches).

Motivated by our design goal of keeping HJ efficient and scalable, we opted for a   greedy heuristic that allows for estimating the impact of evaluating an untested configuration in a light-weight fashion, e.g., avoiding retraining the GP models. More in detail, we simulate the evaluation of an untested configuration $c$ by including $c$ in the set of tested configurations and considering its accuracy to be the one predicted via the GP model's posterior mean. Next, we execute the \textsc{Evaluate\_Jump\_Risk} function to obtain the updated risk of jumping and the corresponding target stage. We repeat this procedure for all the untested configurations in the stage, and select, as the next  to test, the one that enables the longest safest jump (determined via the \textsc{Evaluate\_Jump\_Risk} function). The pseudo-code of this mechanism can be found in the supplemental material.

\subsection{Preserving HyperBand's Robustness}
\label{sec:theo}

An appealing theoretical property of HB is that its exploration policy is guaranteed to be at most a constant factor slower than random search. In order to preserve this property, HJ exploits two  mechanisms: \textbf{i)} with probability $p_{NJ}$ HJ is forced not to jump and abide by the original SH/HB logic; \textbf{ii)} when selecting the configurations to include in a bracket, 
a fraction $p_U$ is selected uniformly at random (as in HB) and not using the model.
The former mechanism ensures that, regardless of how model mispredictions affect HJ's policy, there exists a non-null probability that HJ will not deviate 
from HB's policy (by jumping) in any of the brackets that it executes. The latter mechanism, originally proposed in BOHB~\cite{bohb}, ensures that, independently of the model's behavior, every configuration has a non-null probability of being included in a bracket. We adopt as default value for $p_U$ the same used by BOHB, i.e., 0.3, and use the same value also for $p_{NJ}$.

\subsection{Parallelizing HyperJump}
\label{sec:parallel}


Existing approaches for parallelizing HB can be classified as synchronous or asynchronous. In synchronous approaches~\cite{hyperband,bohb}, the size of the stages abides by HB's rules and parallelization can be achieved  at the level of a HB's stage, bracket, and iteration. Conversely, asynchronous methods (ASHA~\cite{asha} or Ray Tune~\cite{raytune}) consider a single logical bracket that promotes a configuration $c$ to the next stage iff $c$ is in the top $\eta^{-1}$ configurations tested in the current stage.

HJ can be straightforwardly parallelized using synchronous strategies. If parallelization is pursued at the level of brackets or iterations, each worker simply runs an independent instance of HJ that shares the same model and training set (so to share the knowledge acquired by the parallel HJ instances).
When parallelizing the testing of the configurations in the same stage, as soon as a worker completes the testing of a configuration, the model can be updated and the risk of jumping  computed. If a jump is performed, the workers that are still evaluating configurations in the current stage can either be immediately interrupted  or allowed to complete their current evaluation. In our implementation, we opted for the former option, which  has the advantage of maximizing the 
workers that are immediately available for testing configurations in the stage targeted by the jump.

In more detail, our implementation adopts the same parallelization strategy of BOHB, i.e., parallelizing by stage and activating a new parallel bracket in the presence of idle workers, with one exception. We prevent starting a new parallel bracket if there are  idle workers during the first bracket. This, in fact, would reduce the available computational resources to complete the first bracket and, consequently, lead to a likely increase of the latency to recommend the first incumbent (i.e., to test a full budget configuration).

Investigating how to employ HJ in combination with asynchronous  versions of HB is out of the scope of this paper. Yet,  we argue that the key ideas at the basis of HJ could still be applied in this context, opening up an interesting line of future research. For instance, in ASHA, one could use a model-based approach, similar to the one employed by HJ, to: \textbf{(i)} promote 
``prematurely'' configurations that the model predicts to be promising; or \textbf{(ii)} use the model to delay promotions of less promising configurations.

\section{Evaluation}

\label{sec:eval}

This section evaluates HJ both in terms of the quality of the recommended configurations and of its optimization time. We compare HJ against 6 state-of-the-art optimizers using 9 benchmarks and considering both sequential and parallel deployments. 
We also perform an ablation study to dive into the performance of HJ's different components.


\paragraph{Benchmarks.} Our first benchmark is the NATS-Bench~\cite{nats}, which optimizes the topology of a NN, fixing the number of layers and  hyper-parameters, using 2 different data sets (ImageNet-16-120~\cite{imagenet} and  Cifar100). 
The search space encompasses 6 dimensions (each one represents a connection between two layers and have 5 possible topologies).
This benchmark contains an exhaustive evaluation of all possible 15625 configurations.

We also use LIBSVM \cite{libsvm} on the Covertype data set \cite{covertype}, which we sub-sampled by $\approx 5\times$ due to time and hardware constraints. The considered hyper-parameters are the kernel (linear, polynomial, RBF, and sigmoid), $\gamma$, and C. In this case, we could not exhaustively explore off-line the hyper-parameter space, so the optimum is unknown. 

Additional details on the above benchmarks can be found in the supplemental material. Further, the supplemental material reports additional results using 2 alternative data-sets (Cifar10~\cite{cifar10}, 2017 CCF-BDCI~\cite{unet1}, and MNIST~\cite{mnist}) and different models (such as Light UNET~\cite{unet1}, CNN).

\begin{figure*}[t]
    %
    %
    
    \begin{subfigure}[h]{0.33\textwidth}
        \includegraphics[width=\textwidth]{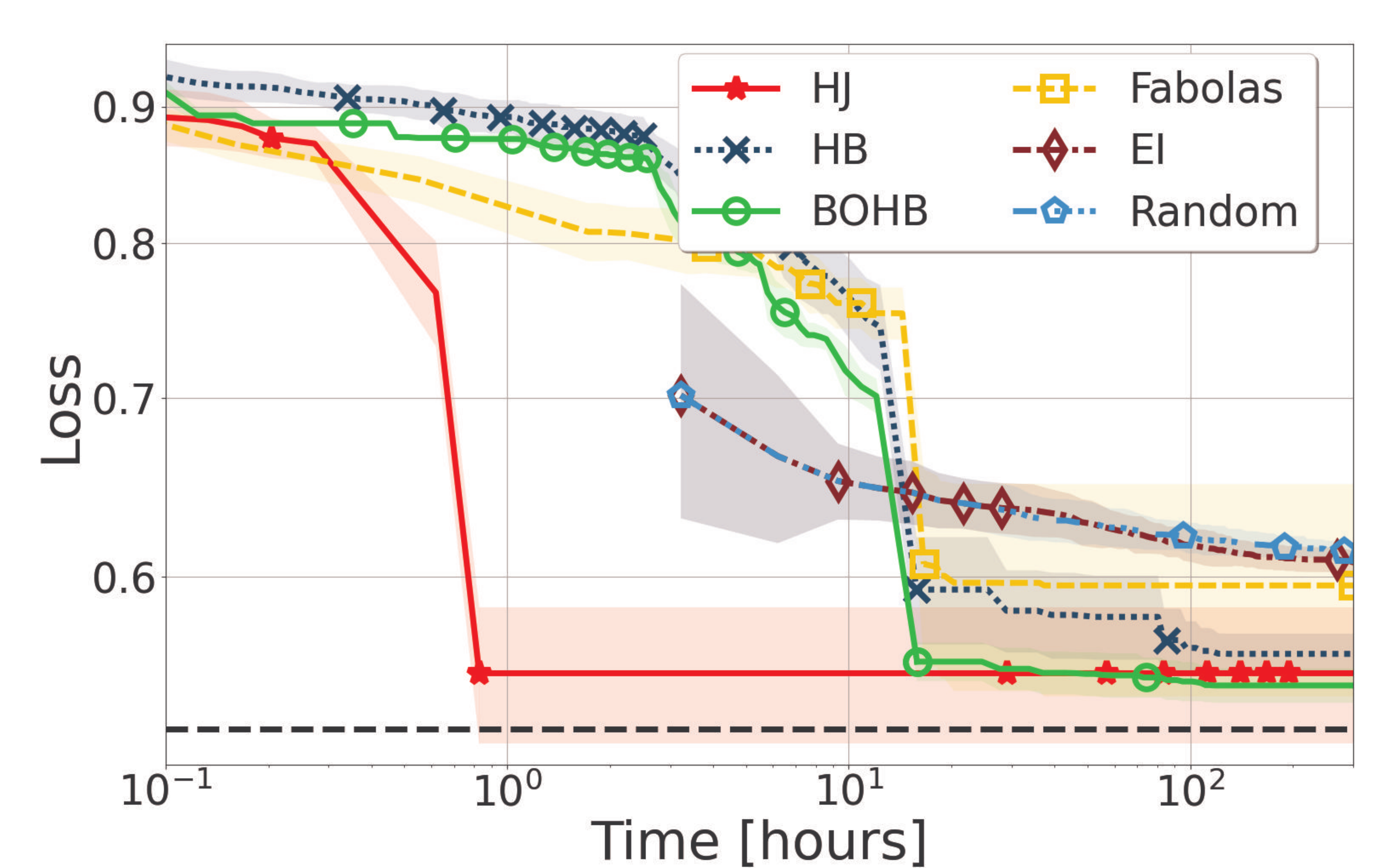}
        \caption{NATS ImageNet (1 worker)}
        \label{fig:opt_nats_imageNet}
    \end{subfigure}    
    \begin{subfigure}[h]{0.33\textwidth}
        \includegraphics[width=\textwidth]{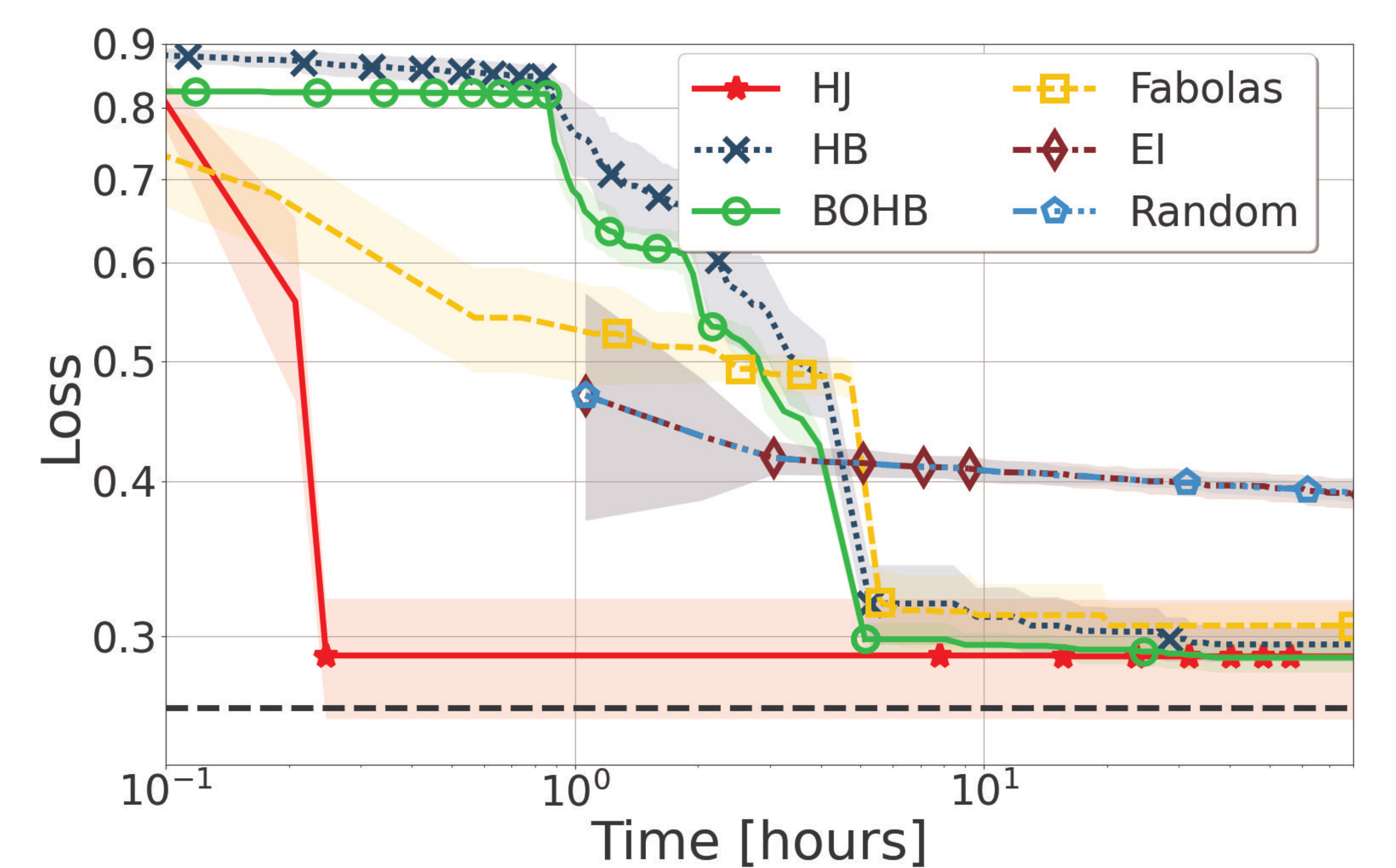}
        \caption{NATS Cifar100 (1 worker)}
        \label{fig:opt_nats_cifar100}
    \end{subfigure}
    %
    %
    \begin{subfigure}[h]{0.33\textwidth}
    \centering
        \includegraphics[width=\textwidth]{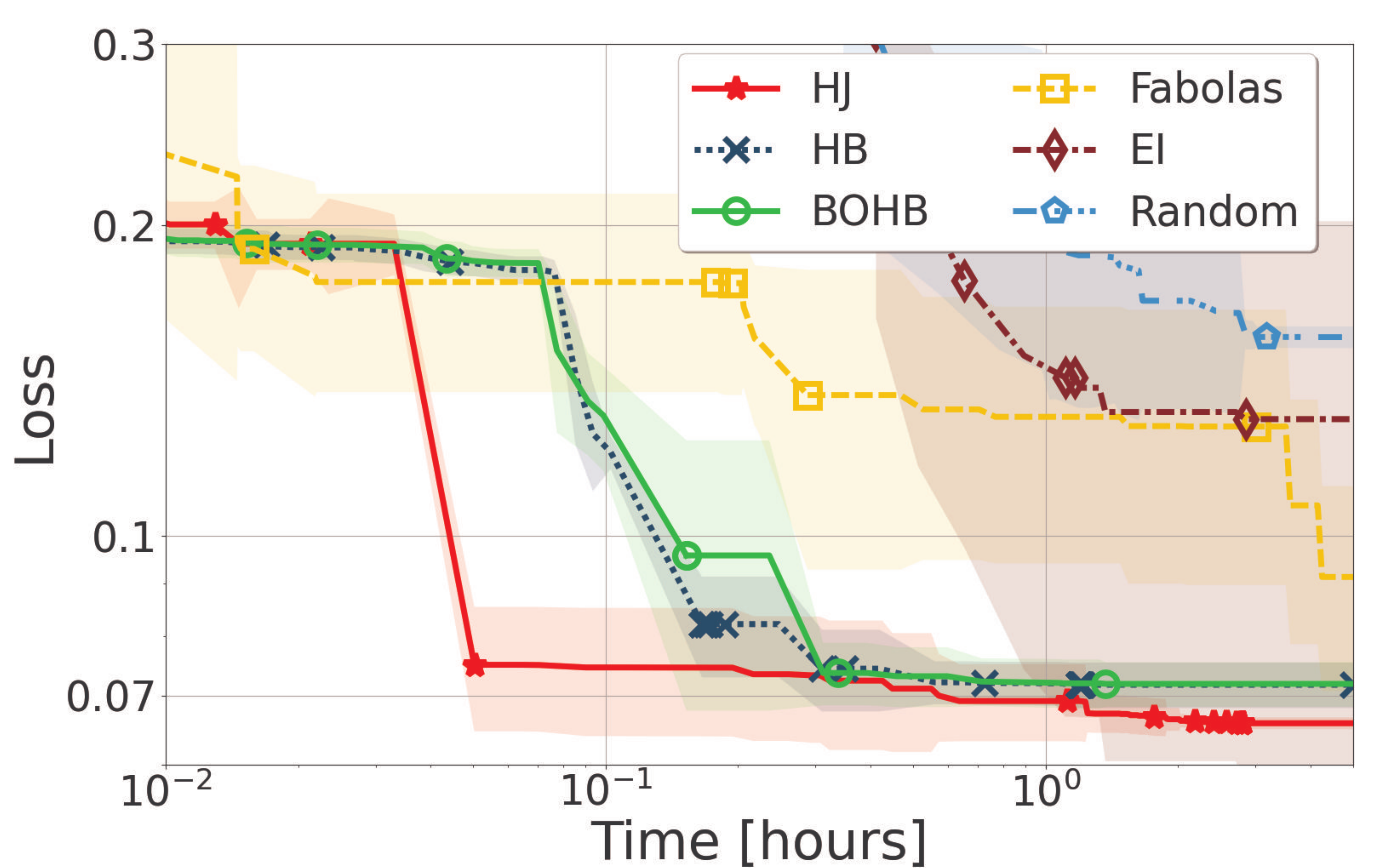}
        \caption{SVM Covertype (1 worker)}
        \label{fig:opt_svm}
    \end{subfigure}
    
    %
    \begin{subfigure}[h]{0.33\textwidth}
        \includegraphics[width=\textwidth]{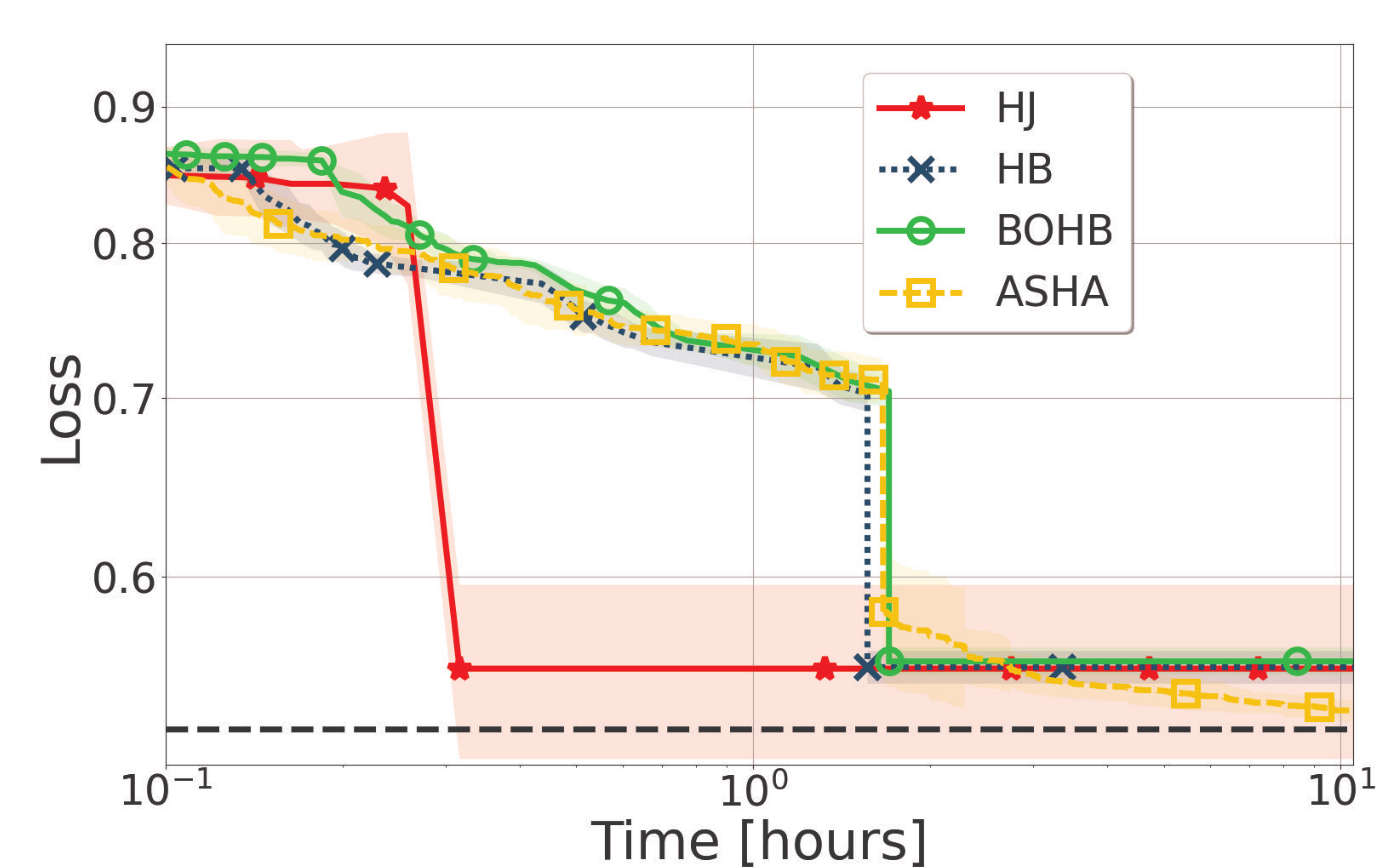}
    \caption{NATS ImageNet (32 workers)}
    \label{fig:parallel_nats_imgNet}
    \end{subfigure}
    \begin{subfigure}[h]{0.33\textwidth}
        \includegraphics[width=\textwidth]{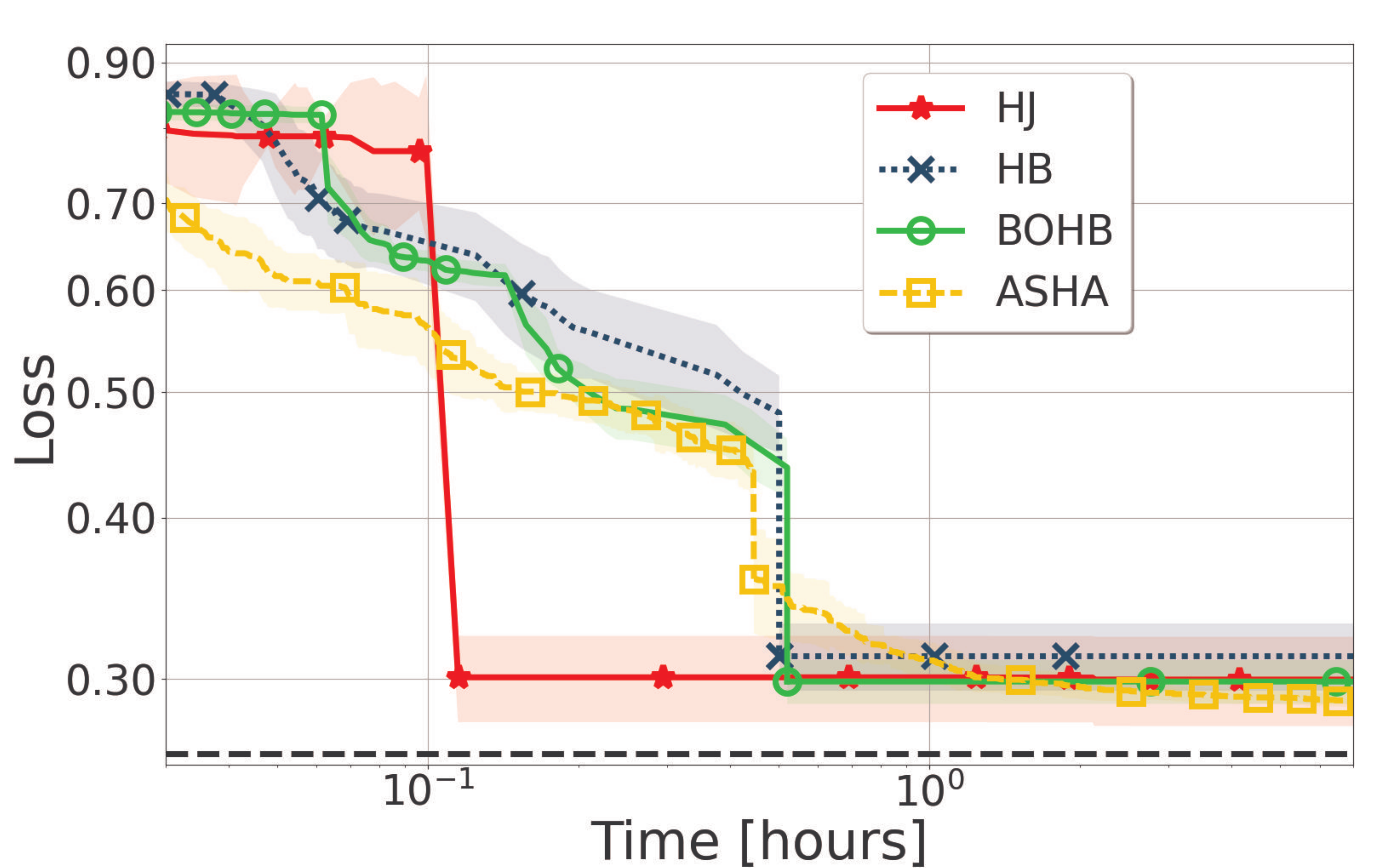}
        \caption{NATS Cifar100 (32 workers)}
        \label{fig:parallel_nats_cifar100}
    \end{subfigure}
    \begin{subfigure}[h]{0.33\textwidth}
        \includegraphics[width=\textwidth]{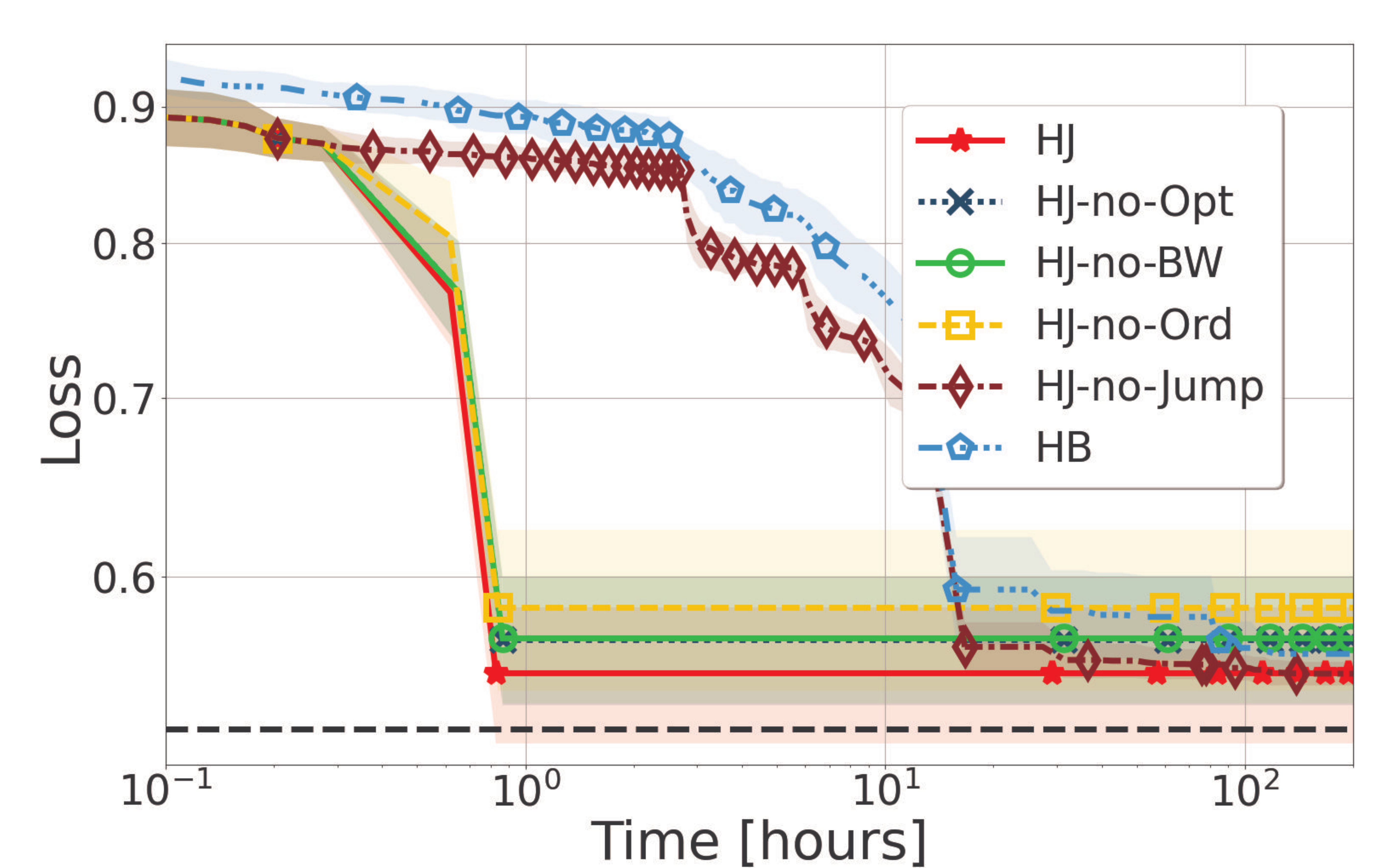}
        \caption{Ablation Study ImageNet (1 worker)}
        \label{fig:abl_nats_imageNet}
    \end{subfigure}    

\caption{Comparison of HJ against other state-of-the-art optimizers in sequential and parallel  deployments. 
Figure (f) reports an ablation study. The dashed horizontal line indicates the optimum (when it is known). 
Additional results are reported in the supplementary material.}
\label{fig:opt}
\end{figure*}


\paragraph{Baselines and experimental setup.}

We compare HJ against six optimizers: HB, BOHB, ASHA, Fabolas, standard BO with EI, and Random Search (RS). The last two techniques (EI and RS) evaluate configurations only with the full-budget.
The implementation of HJ extends the publicly available code of BOHB, which also provides an implementation of HB. ASHA was implemented via the Ray-Tune framework~\cite{raytune}. 
Fabolas was evaluated using its publicly available implementation. 

We use the default parameters of BOHB and Fabolas. 
Similarly to HB, we set the parameter $\eta$  to 3, and for fairness, when comparing HJ, HB, BOHB, and ASHA, we configure them to use the same $\eta$ value.
We use the default value of 10\% for the threshold $\lambda$ for HJ and include in the supplemental material  a  study on the sensitivity to the tuning of $\lambda$. 
We use number of epochs as budget for NATS and training set size for LIBSVM.
The reported results  represent the average of 30 independent runs.
We have made available the implementation of HJ and the  benchmarks used \footnote{\url{https://github.com/pedrogbmendes/HyperJump}}.

\paragraph{Sequential deployment.}
\label{exp_res}
Figure~\ref{fig:opt} reports the average loss (i.e., the test error rate) and corresponding standard deviation in the shaded areas as a function of the wall clock time (i.e., training and recommendation time). 
We start by analyzing the plots in the first rows, which refer to a sequential deployment scenario, i.e., a single worker is available for evaluating configurations. We do not include ASHA in this study, since ASHA is designed for parallel deployments. 

In all the benchmarks, HJ provides significant speed-ups with respect to all the baselines to identify  near optimal configurations. 
The largest speed-ups for recommending near optimal  solutions are achieved with ImageNet and Cifar100, where the gains of HJ w.r.t. the best baseline are around 20$\times$ to 32$\times$, respectively.
Using SVM, the HJ speed-ups to identify good quality configuration (i.e., loss around 0.08) are approx. 7$\times$ w.r.t. to the most competitive baselines, namely HB and BOHB. Note that HJ outperforms these two baselines also in the final stages of the optimization: HJ's final loss is 0.066$\pm$8.9E-4, whereas HB and BOHB's final losses are 0.072$\pm$3.6E-3 and 0.72$\pm$3.5E-3, respectively.

In our benchmarks, BOHB provides marginal (ImageNet and Cifar100) or no (SVM) benefits when compared to HB. 
As shown in the supplemental material (Section~13), this is imputable to the limited accuracy of the modelling approach used by BOHB (based on TPE and on a model per budget) which, albeit fast, is not very effective in identifying high quality configurations to include in a new bracket. Fabolas' performance, conversely, is hindered by its large recommendation times, which are already on the order of a few minutes in the early stages of the optimization and grow more than linearly:  consequently, Fabolas suffers from large overheads especially with benchmarks that have shorter training times, such as SVM. 
Lastly, the limitations of  EI, which only uses full-budget sampling, are notably clear with SVM. Here, training times grow more than linearly with the training set size (which is the budget used by the multi-fidelity optimizers), thus amplifying the speed-ups achievable by using low-fidelity observations w.r.t.~the other benchmarks. 

\paragraph{Ablation study.}
Figure~\ref{fig:abl_nats_imageNet} shows the result of an ablation study aimed at  quantifying the contributions of the various mechanisms employed by HJ.  We report the performance achieved on ImageNet by four HJ variants obtained by disabling each of the following mechanisms: 
 \textbf{(i)} the jumping logic (HJ-no-Jump) (see supplemental material).
\textbf{(ii)} prioritizing the evaluation order of configurations (HJ-no-Ord);  
\textbf{(iii)} the pause-resume training\cite{freeze-thawBO} and opportunistic evaluation  optimizations~\cite{raytune}, see Sec.~6 of the supplemental material  (HJ-no-Opt); 
\textbf{(iv)} the bracket warm-starting~\cite{bohb_similar1} logic, see Sec.~6 of the supplemental material  (HJ-no-BW).

We include in the plot also HB, which can be regarded as a variant of HJ from which we disabled all of the above mechanisms. This data shows that the last two of these mechanisms, which correspond to previously published optimizations of HB, have a similar and small impact on the performance of HJ. 
Conversely, the largest performance penalty is observed when disabling jumping. This confirms that this mechanism is indeed the one that contributes the most to HJ's efficiency. Finally, this plot also shows the relevance of the heuristics that determines the order of configuration testing in a bracket described in Section~\ref{sec:order}.

\paragraph{Parallel Deployment.}
Figures~\ref{fig:parallel_nats_imgNet} and \ref{fig:parallel_nats_cifar100} report the results when using a pool of 32 workers with NATS-Bench with ImageNet and Cifar100. The supplemental material includes the plots for the other benchmarks, as well as for a scenario with 8 workers. 
These data show that HJ achieves gains similar to the ones previously observed compared to HB and BOHB. 
With respect to the other baselines (and to HJ), ASHA adopts a more efficient parallelization scheme (see Section~\ref{sec:parallel}). Thus, the gains of HJ w.r.t. ASHA are slightly reduced. Still, HJ achieves speed-ups of up to approximately 6$\times$ to recommend  near-optimal configurations. This confirms HJ's competitiveness even when compared with HB variants that use optimized parallelization schemes.



\paragraph{Recommendation overhead.} 
We conclude by reporting experimental data regarding the computational overhead incurred by HJ to recommend the next configuration to test. 
The average recommendation time  for HJ  across all benchmarks is approx.~1.08 secs. This time includes model training, determining whether to jump and the next configuration to test in the current stage, and, overall, confirms the computational efficiency of the proposed approach.

\section{Conclusions}
\label{sec:concl}

This paper introduced HyperJump, a new approach that complements HyperBand's robust search strategy and accelerates it by skipping low risk evaluations. HJ's efficiency hinges on the synergistic use of several innovative risk modelling techniques and of a number of pragmatic optimizations. We show that HJ provides over one-order of magnitude speed-ups on a variety of kernel-based and deep learning problems when compared to HB as well as to a number of  state-of-the-art optimizers.


\section*{Acknowledgments}
Support for this research was provided by ANI and Fundação para a Ciência e a Tecnología (Portuguese Foundation for Science and Technology) through the Carnegie Mellon Portugal Program under Grants SFRH/BD/151470/2021 and SFRH/BD/150643/2020, and via projects with references UIDB/50021/2020, 
POCI\mbox{-}01\mbox{-}0247\mbox{-}FEDER\mbox{-}045915, 
C645008882\mbox{-}00000055.PRR,
POCI-01–0247-FEDER-045907, and  
CPCA/A00/7387/2020. 

\bibliography{biblio}

\begin{thebibliography}{40}
\providecommand{\natexlab}[1]{#1}

\bibitem[{Awad et~al.(2021)Awad, Mallik, and Hutter}]{dehb}
Awad, N.~H.; Mallik, N.; and Hutter, F. 2021.
\newblock {DEHB:} Evolutionary Hyberband for Scalable, Robust and Efficient
  Hyperparameter Optimization.
\newblock In Zhou, Z., ed., \emph{Proceedings of the Thirtieth International
  Joint Conference on Artificial Intelligence, {IJCAI} 2021, Virtual Event /
  Montreal, Canada, 19-27 August 2021}, 2147--2153. ijcai.org.

\bibitem[{Bergstra et~al.(2011)Bergstra, Bardenet, Bengio, and K\'{e}gl}]{tpe}
Bergstra, J.; Bardenet, R.; Bengio, Y.; and K\'{e}gl, B. 2011.
\newblock Algorithms for Hyper-Parameter Optimization.
\newblock In \emph{Advances in Neural Information Processing Systems},
  volume~24, 2546--2554. Curran Associates, Inc.

\bibitem[{Bertrand et~al.(2017)Bertrand, Ardon, Perrot, and
  Bloch}]{bohb_similar2}
Bertrand, H.; Ardon, R.; Perrot, M.; and Bloch, I. 2017.
\newblock {Hyperparameter optimization of deep neural networks: combining
  Hperband with Bayesian model selection}.
\newblock In \emph{Proceedings of Conférence sur l’Apprentissage
  Automatique}.

\bibitem[{Breiman(2001)}]{Random_Forest}
Breiman, L. 2001.
\newblock Random Forests.
\newblock \emph{Machine Learning}, 45(1).

\bibitem[{Brochu et~al.(2010)Brochu, Cora, and {de Freitas}}]{boTutorial}
Brochu, E.; Cora, V.~M.; and {de Freitas}, N. 2010.
\newblock A Tutorial on Bayesian Optimization of Expensive Cost Functions, with
  Application to Active User Modeling and Hierarchical Reinforcement Learning.
\newblock Technical Report arXiv:1012.2599.

\bibitem[{Casimiro et~al.(2020)Casimiro, Didona, Romano, Rodrigues, Zwanepoel,
  and Garlan}]{lynceus}
Casimiro, M.; Didona, D.; Romano, P.; Rodrigues, L.; Zwanepoel, W.; and Garlan,
  D. 2020.
\newblock Lynceus: Cost-efficient Tuning and Provisioning of Data Analytic
  Jobs.
\newblock In \emph{Proceedings 20th IEEE International Conference on
  Distributed Computing Systems}.

\bibitem[{Chang and Lin(2011)}]{libsvm}
Chang, C.-C.; and Lin, C.-J. 2011.
\newblock LIBSVM: A Library for Support Vector Machines.
\newblock \emph{ACM Transactions on Intelligent Systems and Technology}, 2.

\bibitem[{Dai et~al.(2019)Dai, Yu, Low, and Jaillet}]{bo_stop}
Dai, Z.; Yu, H.; Low, B. K.~H.; and Jaillet, P. 2019.
\newblock {B}ayesian Optimization Meets {B}ayesian Optimal Stopping.
\newblock In \emph{Proceedings of the 36th International Conference on Machine
  Learning}, volume~97.

\bibitem[{Deng(2012)}]{mnist}
Deng, L. 2012.
\newblock The MNIST database of handwritten digit images for machine learning
  research [Best of the Web].
\newblock In \emph{IEEE Signal Processing Magazine}, volume~29. IEEE.

\bibitem[{Domhan et~al.(2015)Domhan, Springenberg, and Hutter}]{loss_NN_gps}
Domhan, T.; Springenberg, J.~T.; and Hutter, F. 2015.
\newblock Speeding up Automatic Hyperparameter Optimization of Deep Neural
  Networks by Extrapolation of Learning Curves.
\newblock In \emph{Proceedings of the 24th International Joint Conference on
  Artificial Intelligence}.

\bibitem[{Dong et~al.(2021)Dong, Liu, Musial, and Gabrys}]{nats}
Dong, X.; Liu, L.; Musial, K.; and Gabrys, B. 2021.
\newblock {NATS-Bench}: Benchmarking NAS Algorithms for Architecture Topology
  and Size.
\newblock \emph{IEEE Transactions on Pattern Analysis and Machine Intelligence
  (TPAMI)}.
\newblock \mbox{doi}:\url{10.1109/TPAMI.2021.3054824}.

\bibitem[{Dua and Graff(2017)}]{covertype}
Dua, D.; and Graff, C. 2017.
\newblock {UCI} Machine Learning Repository.

\bibitem[{Falkner et~al.(2018)Falkner, Klein, and Hutter}]{bohb}
Falkner, S.; Klein, A.; and Hutter, F. 2018.
\newblock {BOHB}: Robust and Efficient Hyperparameter Optimization at Scale.
\newblock In \emph{Proceedings of the 35th International Conference on Machine
  Learning}, volume~80.

\bibitem[{Golovin et~al.(2017)Golovin, Solnik, Moitra, Kochanski, Karro, and
  Sculley}]{Vizier}
Golovin, D.; Solnik, B.; Moitra, S.; Kochanski, G.; Karro, J.; and Sculley, D.
  2017.
\newblock Google Vizier: A Service for Black-Box Optimization.
\newblock In \emph{Proceedings of the 23rd ACM SIGKDD International Conference
  on Knowledge Discovery and Data Mining}.

\bibitem[{Hutter et~al.(2011)Hutter, Hoos, and Leyton-Brown}]{smac}
Hutter, F.; Hoos, H.~H.; and Leyton-Brown, K. 2011.
\newblock Sequential Model-Based Optimization for General Algorithm
  Configuration.
\newblock In \emph{Proceedings of the 5th International Conference on Learning
  and Intelligent Optimization}, 507–523.

\bibitem[{Jamieson and Talwalkar(2016)}]{successiveHalving}
Jamieson, K.; and Talwalkar, A. 2016.
\newblock Non-stochastic best arm identification and hyperparameter
  optimization.
\newblock In \emph{Proceedings of the 19th International Conference on
  Artificial Intelligence and Statistics}.

\bibitem[{Kandasamy et~al.(2016)Kandasamy, Dasarathy, Oliva, Schneider, and
  Poczos}]{ucb-mf-gp}
Kandasamy, K.; Dasarathy, G.; Oliva, J.~B.; Schneider, J.; and Poczos, B. 2016.
\newblock Gaussian Process Bandit Optimisation with Multi-fidelity Evaluations.
\newblock In Lee, D.; Sugiyama, M.; Luxburg, U.; Guyon, I.; and Garnett, R.,
  eds., \emph{Advances in Neural Information Processing Systems}, volume~29.
  Curran Associates, Inc.

\bibitem[{Klein et~al.(2017)Klein, Falkner, Bartels, Hennig, and
  Hutter}]{fabolas}
Klein, A.; Falkner, S.; Bartels, S.; Hennig, P.; and Hutter, F. 2017.
\newblock Fast Bayesian Optimization of Machine Learning Hyperparameters on
  Large Datasets.
\newblock In \emph{Proceedings of the 20th International Conference on
  Artificial Intelligence and Statistics}, volume~54.

\bibitem[{Klein et~al.(2020)Klein, Tiao, Lienart, Archambeau, and
  Seeger}]{klein2020model}
Klein, A.; Tiao, L.~C.; Lienart, T.; Archambeau, C.; and Seeger, M. 2020.
\newblock Model-based asynchronous hyperparameter and neural architecture
  search.
\newblock \emph{arXiv preprint arXiv:2003.10865}.

\bibitem[{Krizhevsky and Hinton(2009)}]{cifar10}
Krizhevsky, A.; and Hinton, G. 2009.
\newblock Learning multiple layers of features from tiny images.
\newblock Technical report, University of Toronto.

\bibitem[{Lam and Willcox(2017)}]{Lam:2017}
Lam, R.; and Willcox, K. 2017.
\newblock Lookahead Bayesian Optimization with Inequality Constraints.
\newblock In \emph{Proceedings of the 31st International Conference on Neural
  Information Processing Systems}.

\bibitem[{Lam et~al.(2016)Lam, Willcox, and Wolpert}]{lookahed_bo_wilcox}
Lam, R.~R.; Willcox, K.~E.; and Wolpert, D.~H. 2016.
\newblock Bayesian Optimization with a Finite Budget: An Approximate Dynamic
  Programming Approach.
\newblock In \emph{Proceedings of the 29th Neural Information Processing
  Systems Conference}.

\bibitem[{Li et~al.(2018)Li, Jamieson, DeSalvo, Rostamizadeh, and
  Talwalkar}]{hyperband}
Li, L.; Jamieson, K.; DeSalvo, G.; Rostamizadeh, A.; and Talwalkar, A. 2018.
\newblock Hyperband: A novel bandit-based approach to hyperparameter
  optimization.
\newblock \emph{Journal of Machine Learning Research}, 18: 1--52.

\bibitem[{Li et~al.(2020)Li, Jamieson, Rostamizadeh, Gonina, Ben-tzur, Hardt,
  Recht, and Talwalkar}]{asha}
Li, L.; Jamieson, K.; Rostamizadeh, A.; Gonina, E.; Ben-tzur, J.; Hardt, M.;
  Recht, B.; and Talwalkar, A. 2020.
\newblock A System for Massively Parallel Hyperparameter Tuning.
\newblock In Dhillon, I.; Papailiopoulos, D.; and Sze, V., eds.,
  \emph{Proceedings of Machine Learning and Systems}, volume~2, 230--246.

\bibitem[{Liaw et~al.(2018)Liaw, Liang, Nishihara, Moritz, Gonzalez, and
  Stoica}]{raytune}
Liaw, R.; Liang, E.; Nishihara, R.; Moritz, P.; Gonzalez, J.~E.; and Stoica, I.
  2018.
\newblock Tune: A Research Platform for Distributed Model Selection and
  Training.
\newblock \emph{arXiv preprint arXiv:1807.05118}.

\bibitem[{Mendes et~al.(2020)Mendes, Casimiro, Romano, and Garlan}]{trimtuner}
Mendes, P.; Casimiro, M.; Romano, P.; and Garlan, D. 2020.
\newblock TrimTuner: Efficient Optimization of Machine Learning Jobs in the
  Cloud via Sub-Sampling.
\newblock In \emph{2020 28th International Symposium on Modeling, Analysis, and
  Simulation of Computer and Telecommunication Systems}. IEEE.

\bibitem[{Mendes et~al.(2021)Mendes, Casimiro, Romano, and Garlan}]{HJ-arxiv}
Mendes, P.; Casimiro, M.; Romano, P.; and Garlan, D. 2021.
\newblock HyperJump: Accelerating HyperBand via Risk Modelling.
\newblock \emph{CoRR}, abs/2108.02479.

\bibitem[{Mockus et~al.(1978)Mockus, Tiesis, and Zilinskas}]{ei}
Mockus, J.; Tiesis, V.; and Zilinskas, A. 1978.
\newblock The Application of Bayesian Methods for Seeking the Extremum.
\newblock In \emph{Toward Global Optimization}, volume~2, 117--128. Elsevier.

\bibitem[{Poloczek et~al.(2017)Poloczek, Wang, and Frazier}]{miso}
Poloczek, M.; Wang, J.; and Frazier, P. 2017.
\newblock Multi-Information Source Optimization.
\newblock In Guyon, I.; Luxburg, U.~V.; Bengio, S.; Wallach, H.; Fergus, R.;
  Vishwanathan, S.; and Garnett, R., eds., \emph{Advances in Neural Information
  Processing Systems}, volume~30. Curran Associates, Inc.

\bibitem[{Rasmussen and Williams(2006)}]{GP_book}
Rasmussen, C.~E.; and Williams, C.~K. 2006.
\newblock \emph{Gaussian Processes for Machine Learning}.
\newblock Cambridge, MA, USA: MIT Press.

\bibitem[{Ronneberger et~al.(2015)Ronneberger, Fischer, and Brox}]{unet1}
Ronneberger, O.; Fischer, P.; and Brox, T. 2015.
\newblock U-Net: Convolutional Networks for Biomedical Image Segmentation.
\newblock In \emph{Medical Image Computing and Computer-Assisted Intervention
  -- MICCAI 2015}. Springer International Publishing.

\bibitem[{Russakovsky et~al.(2015)Russakovsky, Deng, Su, Krause, Satheesh, Ma,
  Huang, Karpathy, Khosla, Bernstein, Berg, and Fei-Fei}]{imagenet}
Russakovsky, O.; Deng, J.; Su, H.; Krause, J.; Satheesh, S.; Ma, S.; Huang, Z.;
  Karpathy, A.; Khosla, A.; Bernstein, M.; Berg, A.~C.; and Fei-Fei, L. 2015.
\newblock {ImageNet Large Scale Visual Recognition Challenge}.
\newblock \emph{International Journal of Computer Vision (IJCV)}, 115(3):
  211--252.

\bibitem[{Sen et~al.(2018)Sen, Kandasamy, and Shakkottai}]{mf-hp}
Sen, R.; Kandasamy, K.; and Shakkottai, S. 2018.
\newblock Multi-Fidelity Black-Box Optimization with Hierarchical Partitions.
\newblock In Dy, J.; and Krause, A., eds., \emph{Proceedings of the 35th
  International Conference on Machine Learning}, volume~80 of \emph{Proceedings
  of Machine Learning Research}, 4538--4547. PMLR.

\bibitem[{Snoek et~al.(2012)Snoek, Larochelle, and P.~Adams}]{Practical_BO}
Snoek, J.; Larochelle, H.; and P.~Adams, R. 2012.
\newblock Practical Bayesian Optimization of Machine Learning Algorithms.
\newblock In \emph{Proceedings of the 25th International Conference on Neural
  Information Processing Systems}, volume~2.

\bibitem[{Song et~al.(2019)Song, Chen, and Yue}]{general-mfbo}
Song, J.; Chen, Y.; and Yue, Y. 2019.
\newblock A General Framework for Multi-fidelity Bayesian Optimization with
  Gaussian Processes.
\newblock In Chaudhuri, K.; and Sugiyama, M., eds., \emph{Proceedings of the
  Twenty-Second International Conference on Artificial Intelligence and
  Statistics}, volume~89 of \emph{Proceedings of Machine Learning Research},
  3158--3167. PMLR.

\bibitem[{Swersky et~al.(2013)Swersky, Snoek, and Adams}]{mtbo}
Swersky, K.; Snoek, J.; and Adams, R.~P. 2013.
\newblock Multi-task Bayesian Optimization.
\newblock In \emph{Proceedings of the 26th International Conference on Neural
  Information Processing Systems}, volume~2.

\bibitem[{Swersky et~al.(2014)Swersky, Snoek, and Adams}]{freeze-thawBO}
Swersky, K.; Snoek, J.; and Adams, R.~P. 2014.
\newblock Freeze-thaw bayesian optimization.
\newblock \emph{arXiv preprint arXiv:1406.3896}.

\bibitem[{Takeno et~al.(2020)Takeno, Fukuoka, Tsukada, Koyama, Shiga, Takeuchi,
  and Karasuyama}]{mes-par}
Takeno, S.; Fukuoka, H.; Tsukada, Y.; Koyama, T.; Shiga, M.; Takeuchi, I.; and
  Karasuyama, M. 2020.
\newblock Multi-fidelity {B}ayesian Optimization with Max-value Entropy Search
  and its Parallelization.
\newblock In III, H.~D.; and Singh, A., eds., \emph{Proceedings of the 37th
  International Conference on Machine Learning}, volume 119 of
  \emph{Proceedings of Machine Learning Research}, 9334--9345. PMLR.

\bibitem[{Wang et~al.(2018)Wang, Xu, and Wang}]{bohb_similar1}
Wang, J.; Xu, J.; and Wang, X. 2018.
\newblock Combination of Hyperband and Bayesian Optimization for Hyperparameter
  Optimization in Deep Learning.
\newblock \emph{arXiv preprint arXiv:1406.3896}.

\bibitem[{Yue and Kontar(2020)}]{yue2020nonmyopic}
Yue, X.; and Kontar, R.~A. 2020.
\newblock Why Non-myopic Bayesian Optimization is Promising and How Far Should
  We Look-ahead? A Study via Rollout.
\newblock \emph{arXiv preprint arXiv:1911.01004}.

\end{thebibliography}

\clearpage
\includepdf[pages=-]{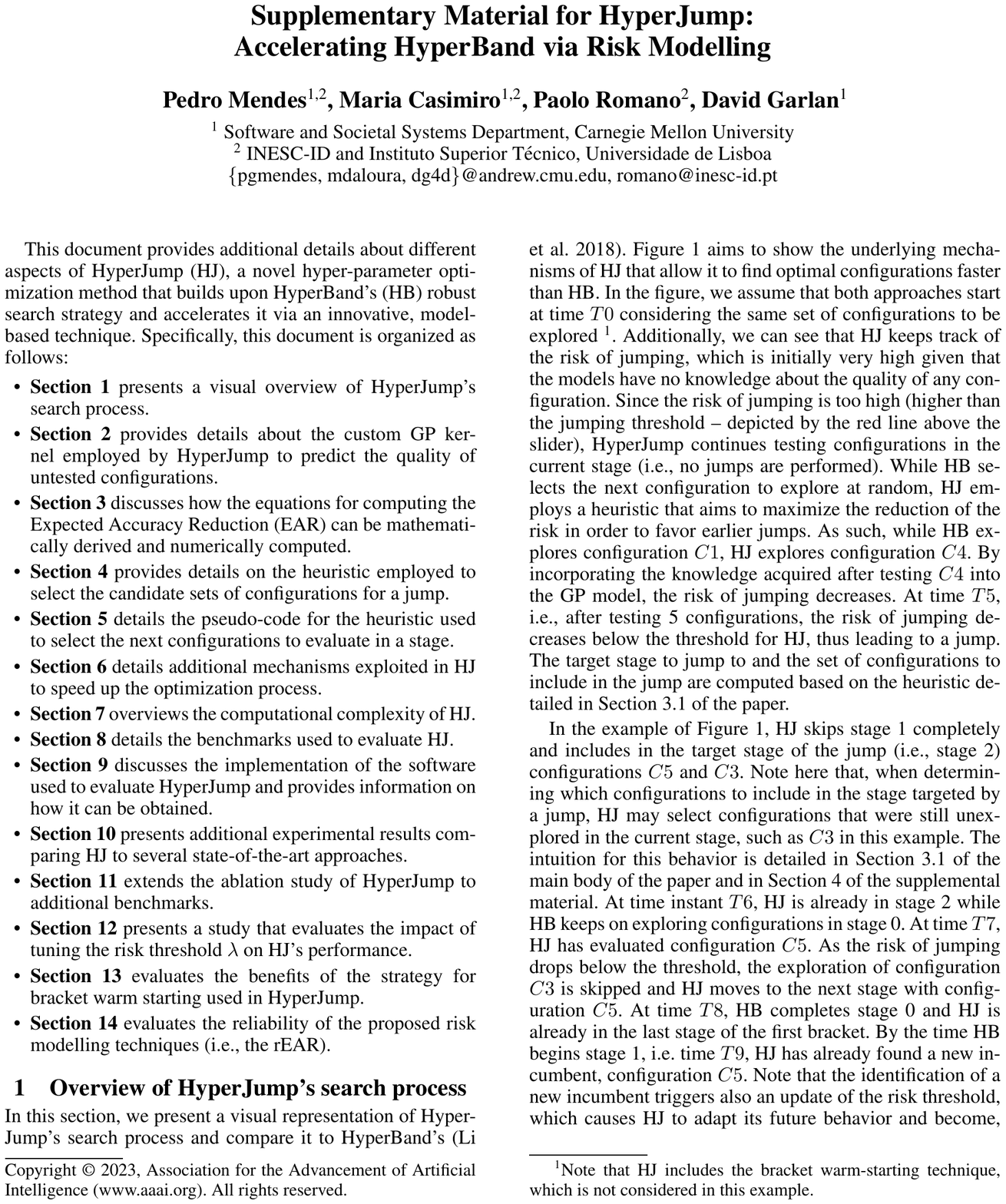}

\end{document}